\documentclass[runningheads]{llncs}
\usepackage{color,soul}
\usepackage[para]{footmisc}
\usepackage{booktabs}
\usepackage{multirow}
\usepackage{graphicx}
\usepackage{amsmath,amssymb,amsfonts}
\usepackage{hyperref}

\begin{document}

\title{Microscopy Image Segmentation via Point and Shape Regularized Data Synthesis}

\author{Shijie Li$^*$\inst{1} \and
Mengwei Ren$^*$\inst{1} \and
Thomas Ach\inst{2} \and
Guido Gerig\inst{1}
}

\titlerunning{Microscopy Image Segmentation via Data Synthesis}
\authorrunning{S. Li and M. Ren et.al}

\institute{Computer Science and Engineering, New York University, Brooklyn, NY, USA. \email{\{shijie.li, mengwei.ren, gerig\}@nyu.edu}  \and
Ophthalmology, University Hospital Bonn, Bonn, Germany}
\maketitle              
\def\thefootnote{*}\footnotetext{These authors contributed equally to this work}\def\thefootnote{\arabic{footnote}}
\begin{abstract}
Current deep learning-based approaches for the segmentation of microscopy images heavily rely on large amount of training data with dense annotation, which is highly costly and laborious in practice. Compared to full annotation where the complete contour of objects is depicted, point annotations, specifically object centroids, are much easier to acquire and still provide crucial information about the objects for subsequent segmentation. In this paper, we assume access to point annotations only during training and develop a unified pipeline for microscopy image segmentation using synthetically generated training data. 
Our framework includes three stages: (1) it takes point annotations and samples a pseudo dense segmentation mask constrained with shape priors; (2) with an image generative model trained in an unpaired manner, it translates the mask to a realistic microscopy image regularized by object level consistency; 
(3) the pseudo masks along with the synthetic images then constitute a pairwise dataset for training an ad-hoc segmentation model. 
On the public MoNuSeg dataset, our synthesis pipeline produces more diverse and realistic images than baseline models while maintaining high coherence between input masks and generated images. When using the identical segmentation backbones, the models trained on our synthetic dataset significantly outperform those trained with pseudo-labels or baseline-generated images. Moreover, our framework achieves comparable results to models trained on authentic microscopy images with dense labels, demonstrating its potential as a reliable and highly efficient alternative to labor-intensive manual pixel-wise annotations in microscopy image segmentation. The code can be accessed through \url{https://github.com/CJLee94/Points2Image}. 

\end{abstract}

\section{Introduction} 
Instance segmentation is vital in medical imaging as it accurately segments individual objects of interest, allowing clinicians to explore the statistical pattern of diseases. Although learning-based methods have excelled, the lack of large-scale datasets with dense annotations is a significant challenge. Manually outlining each object in closely clustered 2D microscopy images is time-consuming and laborious, requiring extensive domain expertise. This problem is even more pronounced in 3D acquisitions.

To address this limitation, various approaches have been adopted based on unsupervised pretraining coupled with one/few-shot finetuning~\cite{ciga2022self}
which have shown promise in reducing the number of required annotations. However, acquiring accurately annotated samples for the finetuning is still extremely challenging. 
Another microscopy image labeling strategy that reduces costs  is to perform sparse annotation such as bounding boxes, scribbles~\cite{qu2020weakly}, or center points.
These sparse annotations can be utilized to generate a pseudo mask or probability map~\cite{tian2020weakly,liu2022weakly,li2021point,chen2022unsupervised} to train the segmentation model. However, compared to fully supervised methods, they may compromise accuracy and fail on objects with fuzzy boundaries or overlapping structures.

Alternatively, training on synthesized data is another option to compensate for insufficient training datasets by artificially generating large numbers of image and segmentation pairs. 
For instance,~\cite{billot2021synthseg,dey2023anystar} sample images with randomized contrast and resolution from an input label map, greatly benefiting biomedical segmentation.
~\cite{hou2017unsupervised,hou2019robust} generate microscopy patches by separately synthesizing foreground/background textures. Yet, they require color-based super segmentation, thus limiting their generalization ability to other image modalities.
Conditional GANs~\cite{pix2pix2017,CycleGAN2017} have been employed for semantic image synthesis, which generate realistic images conditioned on an input semantic label~\cite{park2019SPADE,park2019semantic,sushko2020you}. 
Accordingly, variants of these semantic synthesis models have then been developed for mask-to-microscopy image translation~\cite{falahkheirkhah2023deepfake,butte2023enhanced,wang2022sian,butte2022sharp,lou2022pixel} to enhance the microscopy image analysis. Nevertheless, they still rely on pairwise dense label and image during the training stage.
Conversely, unpaired image-to-image translation methods~\cite{huang2018munit,CycleGAN2017} do not rely on pairwise data, and are repurposed for microscopy images synthesis~\cite{mahmood2019deep,fu2018three,liu2021asist}. However, ~\cite{mahmood2019deep} requires a joint training using both synthetic and real pairs which is infeasible in our setup as we only have access to point annotations.
The most relevant works~\cite{fu2018three,liu2021asist} apply CycleGAN to 3D grayscale microscopy dataset, enabling segmentation without groundtruth labels. However, naive re-purposing of CycleGAN without carefully-designed regularizations may lead to inferior performance in color microscopy images where the background patterns are much more complicated than in grayscale images. In our empirical experiments, we observe issues such as contrast inversion between foreground and background, as well as limited image diversity (see Fig.~\ref{fig:synthesis_results}).

In this paper, we propose a novel unified microscopy image synthesis and segmentation pipeline that uses only point annotation and image pairs during training. 
In particular, the synthesis component takes a point label as the input, then transforms the points into a stochastic shape-regularized instance mask, and finally  creates a point-regularized microscopy image.  
Once trained, our synthesis pipeline can generate a diverse set of image-segmentation pairs from a set of real/ synthetic point labels. The synthetic data is used for training a segmentation backbone under supervised learning objectives.
Our contributions are threefold. First, we introduce a unified microscopy synthesis and segmentation  framework that  requires training images with only point annotations. Secondly, we incorporate stochasticity and regularizations into the synthesis pipeline to enhance the diversity of the synthetic dataset while maintaining object-consistency between the generated instance masks and synthetic images. 
Finally, we train a segmentation network using the synthetic dataset and demonstrate superior performance compared to existing benchmarks at both pixel and object levels. Our results are comparable to those obtained using training on real images with dense annotations, which indicates the potential of our approach to significantly reduce annotation costs.

\section{Methods}
\begin{figure}[t]
    \centering
    \includegraphics[width=0.9\textwidth]{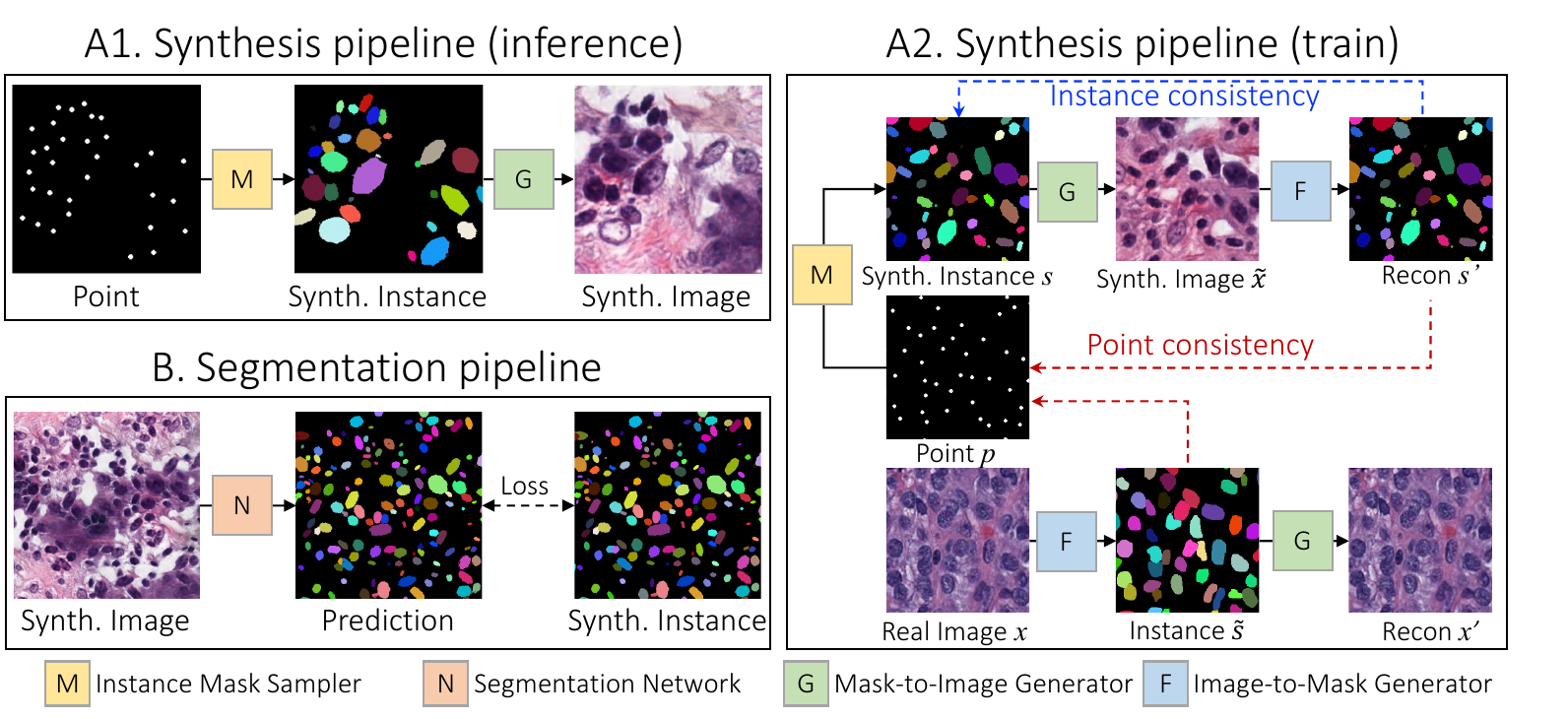}
    \caption{Framework overview. A1/A2 illustrate the inference/training of our synthesis pipeline that takes a point label as input and samples an instance/segmentation pair. B. is the segmentation backbone trained with the synthetic data in a supervised fashion.
    }
    \label{fig:synthesis_pipeline}
\end{figure}
 
\noindent\textbf{Formulation.} Fig.~\ref{fig:synthesis_pipeline} gives an overview of our framework consisting of \textbf{A}. synthesis and \textbf{B}. segmentation components. 
The synthesis pipeline samples an instance mask given the input point label with shape priors, and translates the instance mask to a microscopy image. After obtaining a set of point annotations, a synthetic dataset is generated with both synthetic images and instance mask pairs. The point annotations can be either manually created or randomly generated. This dataset is then used to train the segmentation model B.

Formally, we assume access during training to a set of microscopy images $X=\{x_1, x_2, \dots, x_n\}\in \mathcal{X}$ along with their point annotations $P=\{p_1, p_2,\dots, p_n\}\in \mathcal{P}$. The notation $\mathcal{X}$, $\mathcal{S}$, $\mathcal{P}$ stands for image space, segmentation space, and point mask space separately. 
The instance mask sampler $M:\mathcal{P}\rightarrow \mathcal{S}$ takes the point labels and samples a set of stochastic instance masks $S=\{s_1, s_2, \dots, s_n\}$, where $s_i = M(p_i)$.
Then, a conditional generator $G: \mathcal{S} \rightarrow \mathcal{X}$ translates the masks to synthetic microscopy images $\tilde{X}=\{\tilde{x}_1, \tilde{x}_2, \dots, \tilde{x}_n\}\in \mathcal{X}$, $\tilde{x}_i=G(s_i)$, trained in an unpaired manner as in CycleGAN, which involves a reverse mapping $F:\mathcal{X}\rightarrow \mathcal{S}$ (Fig.~\ref{fig:synthesis_pipeline} A2).
$(\tilde{X},S)$ then constitute a synthetic pairwise dataset that can be used to train the segmentation backbone $N$. 

\noindent\textbf{Shape-regularized instance mask sampler.}
We use shape distribution priors to generate a pseudo instance mask from a given input point label map.  
The generation process ($M$ in Fig.~\ref{fig:synthesis_pipeline} A) takes the object centroids as input and it randomly generates an ellipse from each point using a sampled area $\mathcal{A}$, angle $\theta$, and eccentricity $\epsilon$. The area $\mathcal{A}$ is sampled from a uniform distribution $\mathcal{U}(a,b(\rho))$, where $a$ is the lower bound of the object area, and $b$ is the upper-bound depends inversely on the object density $\rho$., which is equivalent to the distance to the nearest point. This prevents severe overlapping between objects in densely crowded regions.\\

\noindent\textbf{Point-regularized mask-to-image synthesis.}
\label{sec:GAN}
To train the mask-to-image generator $G$ without access to mask-image pairs, we construct $G$ under the commonly used unpaired image translation framework CycleGAN\cite{CycleGAN2017}. As shown in Fig.~\ref{fig:synthesis_pipeline} A2, $G$ maps the instance mask $S$ to an image $X$ so that $G(S)$ is indistinguishable to an image discriminator $D_X$ (we omit $D_X$ and $D_S$ in Fig.~\ref{fig:synthesis_pipeline} for simplicity). In practice, we represent $S$ as a concatenation of the semantic mask and the horizontal/vertical maps~\cite{graham2019hover}. 
Additionally, an inverse mapping $F$ enforces the cycle consistency $F(G(S)) \approx S$. To ensure that the output distribution of $F$ aligns with our expectations (i.e., a segmentation map), we formulated $F$ using an instance segmentation framework \cite{graham2019hover}, where $F=\{F_\text{hv},F_\text{seg}\}$ predicts both the horizontal/vertical distance maps and the semantic mask. Consequently, we replaced its cycle consistency loss with an instance segmentation loss (Eq. \ref{eq:hover}). We further incorporate two important considerations into the CycleGAN framework to improve the synthesis quality and fidelity. \\

\noindent\textit{\underline{Diversity}}. 
As the domains of the image and the instance mask possess different complexity (i.e., the mapping between a mask and an image is one-to-many), it contradicts the one-to-one mapping in CycleGAN~\cite{chu2017cyclegan}. Empirically, we find this formulation leads to low diversity of the generated images, particularly within the background (see Fig.~\ref{fig:synthesis_results}a). We speculate using such synthetic datasets may yield detrimental segmentation outcomes.
To this end, we inject a 3D noise tensor at every layer of the generator (including the input layer) to enable the multimodal synthesis that varies in both local and global image appearances. We also replaced the ResNet generator with the OASIS generator~\cite{sushko2020you} where the network architecture is specifically tailored for semantic image synthesis. \\

\noindent\textit{\underline{Point regularization}}. While incorporating randomness allows for the generation of more diverse samples, it may also indicate a less tractable manipulation of the image content, such as inverting the image color and/or contrast~\cite{cohen2018distribution,zhang2019harmonic}. 
In our application, this phenomenon manifests as a color and texture inversion occurring between the foreground and background elements, as illustrated in Fig. \ref{fig:synthesis_results}b, as some organs in the background have a similar elliptical appearance, which can confuse the model if no regularizer is used. A more regularized process~\cite{ren2021segmentation} is thus necessary to ensure that the generated image and the input instance mask exhibit a high level of object-level consistency, while adhering to the characteristics of the underlying imaging modality.
To do so, we employ the point label $P$ shared by both $X$ and $G(S)$ as a foreground indicator to regularize the output of $F$. In particular, applying the point regularizer to $F(X)$ enforces $F$ to accurately identify foreground and background pixels. In turn, regularization on $F(G(S))$ guarantees that the synthetic image accurately captures the color distribution of objects.

\noindent\textit{\underline{Objectives}}. The framework is trained end-to-end with the CycleGAN adversarial terms $\mathcal{L}_{GAN}$, the cycle-consistency terms, as well as the point-consistency terms. We employ LSGAN~\cite{mao2017least} for the adversarial term as:
\begin{align}
    \footnotesize
    \begin{split}
    \mathcal{L}_{\mathrm{GAN}}(D_X) &= \frac{1}{2}\mathbb{E}_{x \sim X}\left[||1-D_X(x)||^2_2\right] + \frac{1}{2}\mathbb{E}_{s \sim S}\left[||D_X(G(s))||^2_2\right],\\
    \mathcal{L}_{\mathrm{GAN}}(G) &= \frac{1}{2}\mathbb{E}_{s\sim S}\left[||D_X(G(s)) -1 ||_2^2\right].
    \end{split}
\end{align}
Analogous terms are used for $\mathcal{L}_{\mathrm{GAN}}(D_S,F)$. 
To enforce cycle consistency, 
we use L1 loss between $G(F(X))$ and $X$, but replace the consistency between $F(G(I))$ and $I$ to an instance-specific loss~\cite{graham2019hover} to accommodate for different domains of X (image) and S (segmentation) defined as follows: 
\begin{align}
\label{eq:hover}
    \footnotesize
    \begin{split}
        \mathcal{L}_{\text{instance}}(F,G) &= \mathbb{E}_{s\sim S}\left[\mathcal{L}_{\text{hv}}(F_{hv},G) + \mathcal{L}_{\text{seg}}(F_{seg},G)\right],\\
        \mathcal{L}_{\text{hv}}(F_{hv},G) &= \mathbb{E}_{s\sim S}\left[||F_{\text{hv}}(G(s))-s_{\text{hv}}||^2_2 + ||\nabla F_{\text{hv}}(G(s))-\nabla s_{\text{hv}}||^2_2\right],\\
        \mathcal{L}_{\text{seg}}(F_{seg},G)&=\mathbb{E}_{s\sim S}\left[Dice(F_{\text{seg}}(G(s)), i_{seg})+CE(F_{\text{seg}}(G(s)), s_{seg})\right].
    \end{split}
\end{align}
 $F_\text{hv}$ predicts the horizontal and vertical distance maps from the generated image, and $F_\text{seg}$ predicts the binary semantic mask.
The point-consistency loss between $F(G(S))$ and $P$, and $F(S)$ and $P$ is defined as 
\begin{align}
    \footnotesize
    \mathcal{L}_{\text{point}}(F,G)= \mathbb{E}_{(s,x,p)\sim (S,X,P)} \left[ -p\log F(G(s)) - p\log F(x)\right].
\label{eq:point_reg}
\end{align}
$s$ and $x$ by definition share the identical point label $p$, as shown in Fig.~\ref{fig:synthesis_pipeline} A2.\\

\noindent\textbf{Segmentation with synthetic training data.}
With the trained synthesis pipeline, we simulate a dataset from a set of point labels which is then used to train an instance segmentation model $N$ with supervised objectives. In practice, $N$ can be instantiated with different network backbones such as~\cite{graham2019hover,kumar2017dataset}.

\section{Experiments}
\noindent{\textbf{Setup and metrics.}}   
We conduct experiments on the MoNuSeg dataset~\cite{kumar2017dataset,kumar2019multi}. A total of 37 2D microscopy images (size $1000\times1000$) obtained from 9 tissues are provided, and 14 testing images are held-out for evaluating the trained segmentation models. 
We use FID (Fréchet Inception Distance) and KID (Kernel Inception Distance)~\cite{binkowski2018demystifying} to measure the visual fidelity of the generated images. We also calculate the standard deviation of 10 samples to measure the diversity of the generated images for the same input instance mask.
Segmentation performance on the test set is quantified by pixel-level Intersection over Union (IoU) and F1 scores as well as by object-level Dice and Average Jaccard Index (AJI). \\

\noindent{\textbf{Implementation details.}} 
\label{sec:implementation}
All models are implemented in PyTorch 1.8.1 and trained with Adam optimizer 
on a single NVIDIA RTX8000 GPU. 
Training takes around 12 hours for the synthesis pipeline and 4 hours for the segmentation pipeline. 
More training/inference details are provided in the appendix. \\

\begin{figure}[t]
    \centering
    \includegraphics[width=\textwidth]{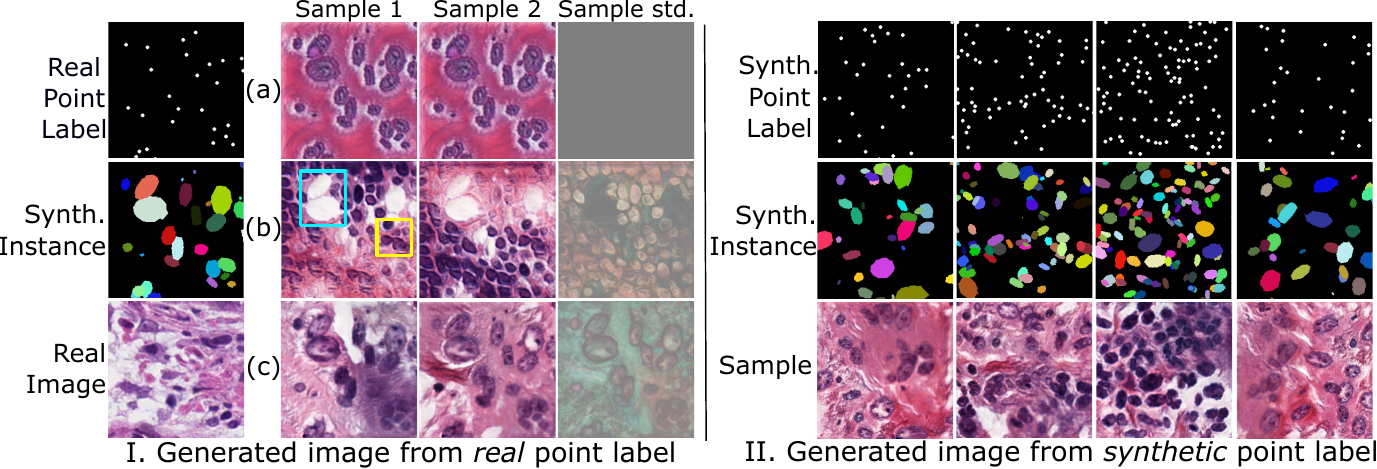}
    \caption{I. Synthetic samples with a \textit{real} point label from the test set. \textbf{(a)} Use of CycleGAN~\cite{CycleGAN2017} lacks texture diversity. ~\textbf{(b)} Injection of  randomness to baseline results in increased diversity but generates incoherent objects and foreground/background color inversion. Our model \textbf{(c)} imposes object/point regularization and produces diverse and proper object samples.
II. Synthetic samples from fully \textit{synthetic} point labels. Our pipeline generates diverse instance masks and anatomically consistent images.}
    \label{fig:synthesis_results}
\end{figure}
\begin{figure}[t]
    \centering
    \includegraphics[width=0.85\textwidth]{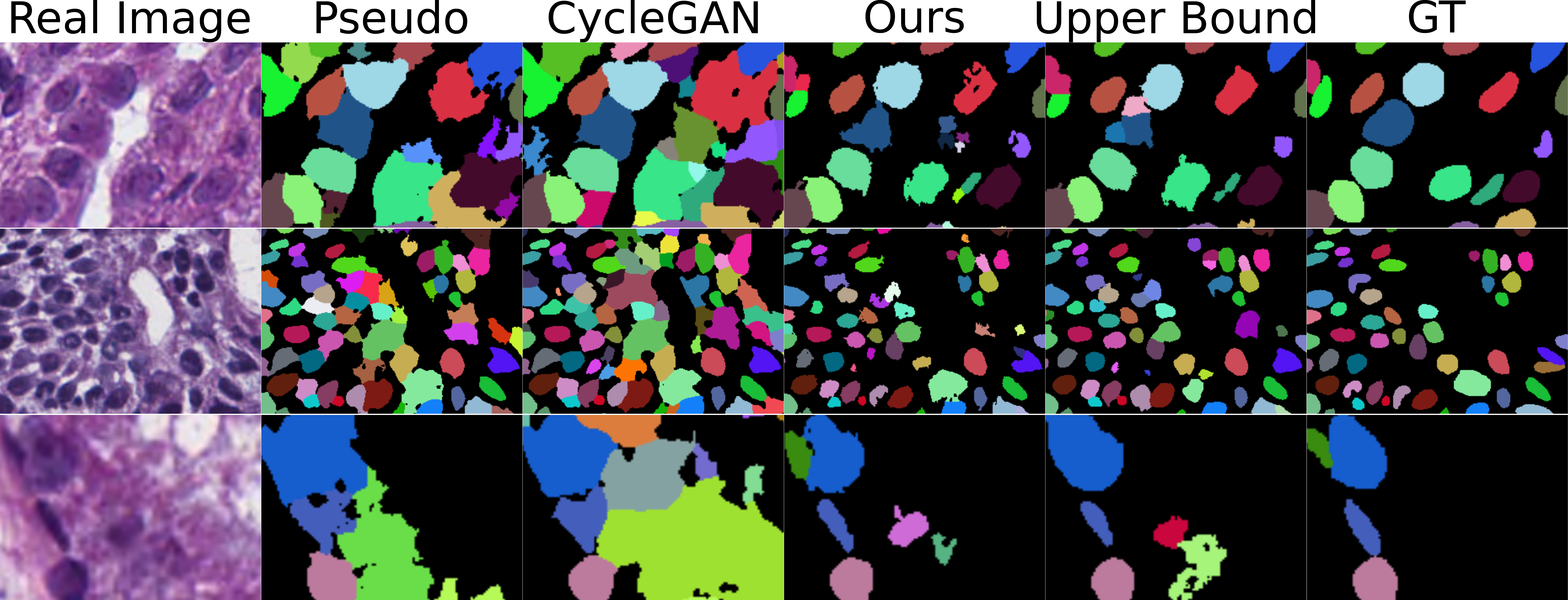}
    \caption{Comparison of instance segmentation results using an identical segmentation backbone \cite{graham2019hover} trained on different data pair generation strategies. Models trained with our synthetic data outperforms models trained on pseudo labels or CycleGAN-generated data pairs, approaching the accuracy of models trained on real images with dense annotation. Challenging regions (highlighted by cyan boxes) with fuzzy boundaries are highlighted. More full-slice results are provided in the appendix.}
    \label{fig:segmentation_result}
\end{figure}
\begin{table}[b]
\centering
\footnotesize
\setlength{\tabcolsep}{5pt}
\begin{tabular}{cccccc}
\toprule               
Config.                   & 
Randomness               & 
Point Reg.                &
FID $\downarrow$          & 
KID $\downarrow$          & 
Diversity $\uparrow$     \\ \hline

(a)      & -              & - 
& 215.58(1.274) & 0.235(0.003) & 4.34e-08 \\
(b)      & \checkmark     & -  
& 213.45(2.273) & 0.189(0.002) & 0.085    \\
(c)      & \checkmark     & \checkmark 
& \textbf{102.44(0.767)} & \textbf{0.072(0.002)} & \textbf{0.166}   \\
\bottomrule
\end{tabular}
\caption{Ablation comparison of image generation quality using inception scores FID, KID, and sample diversity. (a) is a baseline CycleGAN model~\cite{CycleGAN2017} lacking sample diversity. (b) injects noise, reducing inception distances and increasing diversity, but with erroneous object manipulation (see Fig.~\ref{fig:synthesis_results}(b)). Our final model (c) achieves the best inception scores and increased sample diversity.}
\label{tab:inception}
\end{table}
\begin{table}[t]
\footnotesize
\setlength{\tabcolsep}{0.7pt}
\centering
\resizebox{0.95\columnwidth}{!}{
\begin{tabular}{llllcccc}
\toprule
\multirow{2}{*}{Backbone} & 
\multirow{2}{*}{Method} &
\multirow{2}{*}{Image} & 
\multirow{2}{*}{Label} &
\multicolumn{2}{c}{Pixel level} 
& \multicolumn{2}{c}{Object level} \\\cmidrule(lr){5-6} \cmidrule(lr){7-8}
& &                          &                         & IoU $\uparrow$  & F1 $\uparrow$ & Dice $\uparrow$ & AJI $\uparrow$ \\ \midrule
                          & Up.bound        & Real & Dense        
                          & 0.614(0.05)    & 0.760(0.03)  & 0.734(0.03)      & 0.541(0.05)   \\ 
CNN3        & Pseudo~\cite{li2021point} & Real & Point
                          & 0.508(0.08)    & 0.670(0.06)  & 0.587(0.08)      & 0.311(0.11)   \\
~\cite{kumar2017dataset}        & CycleGAN~\cite{CycleGAN2017} & Syn & Point
                          & 0.469(0.11)    & 0.631(0.10)  & 0.547(0.10)    & 0.266(0.12)   \\
        & Ours            & Syn  & Point
                          & \textbf{0.600(0.05)}    & \textbf{0.749(0.04)}  & \textbf{0.691(0.04)}    & \textbf{0.463(0.06)}   \\ \midrule
 & Up.bound       & Real & Dense  
                          & 0.644(0.07)   & 0.781(0.01) & 0.747(0.05)   & 0.572(0.07)  \\ 
HoverNet         &Pseudo~\cite{li2021point} & Real & Point
                          & 0.538(0.07)   & 0.697(0.06) & 0.679(0.05)   & 0.472(0.06)  \\
~\cite{graham2019hover}  &CycleGAN~\cite{CycleGAN2017} & Syn & Point
                          & 0.483(0.19)   & 0.622(0.23) & 0.601(0.20)   & 0.413(0.17)  \\
         & Ours           & Syn & Point
                          & \textbf{0.636(0.05) }  & \textbf{0.776(0.04)} & \textbf{0.745(0.04) }  & \textbf{0.569(0.05)}   \\ \bottomrule
\end{tabular}}
\caption{Segmentation results with various backbones. Models using our synthetic data outperform those trained on pseudo labels or using CycleGAN-generated data, approaching the accuracy of models trained on densly annotated real images (Up.bound).}
\label{tab:benchmark}
\end{table}

\noindent\textbf{Microscopy Image Synthesis.}
We studied the effects of introducing randomness and point regularization on the quality and diversity of samples generated by isolating each components with three different model configurations (Table~\ref{tab:inception} and Fig.~\ref{fig:synthesis_results}): Configuration (a) indicates a baseline CycleGAN; 
configuration (b) introduces randomness into (a), and (c) is our model with further object regularization.
In Fig.~\ref{fig:synthesis_results} (a), CycleGAN samples lacked diversity and visual fidelity, as also quantified by the low standard deviation and high FID/KID scores in Table~\ref{tab:inception}. Introducing randomness in (b) improved both FID and KID and increased sample diversity. However, we observed noticeable color and texture inversion between foreground/background semantics where the colors were incorrectly learned (Fig.~\ref{fig:synthesis_results} (b)). Moreover, intractable generation of non-existing objects appeared in the background. 
Our method (c) with regularization generated visually more realistic samples, and exhibited much higher object-level consistency with the input instance mask. In what follows, we use the generator trained with config (c) for data synthesis and ad-hoc segmentation training.\\

\noindent\textbf{Segmentation Results.} We evaluate the usability of our synthetic data by training a segmentation model using the generated image-segmentation pairs.\\
\noindent\textit{\underline{Benchmark}} We trained the same segmentation backbone under different data configurations: (1) real images with dense manual annotation which serves as the upper bound for all methods `Upper bound'; (2) real images with manual point annotation~\cite{li2021point} `Pseudo', where pseudo dense labels are generated from the point annotation; (3) `CycleGAN'~\cite{CycleGAN2017} and (4) `Ours' following the pipeline in Fig.~\ref{fig:synthesis_pipeline} that artificially synthesize the pairwise data given the real point annotations. 
Table~\ref{tab:benchmark} presents pixel/object-level results on the held-out test set. When only point labels are accessible (our assumption), a large performance gap is observed with pseudo labels compared with the upper bound. 
Training with CycleGAN synthesized dataset does not generalize well to the real test set, indicating a potential lack of data diversity and fidelity as also quantified in Table.~\ref{tab:inception} (a). 
With the same amount of generated data from our pipeline, we achieved highly comparable and/or on-par performance with the upper bound while using only point annotations for training.\\ 

\begin{table}[t]
\centering
\footnotesize
\setlength{\tabcolsep}{6pt}
\resizebox{0.95\columnwidth}{!}{
\begin{tabular}{ccccc}
\toprule
\multirow{2}{*}{\# Synthetic pairs} & 
\multicolumn{2}{c}{Pixel level} & 
\multicolumn{2}{c}{Object level}            \\ 
\cmidrule(lr){2-3} \cmidrule(lr){4-5}
          & IoU $\uparrow$       & F1 $\uparrow$        & Dice $\uparrow$      & AJI $\uparrow$       \\ \midrule
                          23              & 0.629(0.046)         & 0.771(0.035)         & 0.718(0.042)         & 0.530(0.055)         \\
                           46              & 0.633(0.045)         & 0.775(0.034)         & 0.728(0.035)         & 0.544(0.047)         \\
                          69             & 0.640(0.045)         & 0.780(0.036)         & 0.741(0.036) & 0.565(0.047) \\
                          92             & \textbf{0.647(0.047)} & \textbf{0.785(0.036)} & \textbf{0.752(0.039)} & \textbf{0.580(0.051)} \\ \bottomrule
\end{tabular}}
\caption{Segmentation results trained with fully synthetic training data. We train the same backbone (HoverNet) under increasing numbers of generated pairs and observe performance gains at both pixel and object level. Under 92 fully synthetic pairs, the performance surpasses the model trained with 23 real images with dense annotation (see `Up. bound' under HoverNet in Table~\ref{tab:benchmark}). 
\label{tab:datasize}
}
\end{table}

\noindent\textit{\underline{Training with fully synthetic dataset}}. 
We investigate the performance of a fully synthetic dataset, excluding any real point labels during the data generation process. We randomly sample a set of point labels based on empirical assumptions about the object distribution per image and apply the data generation pipeline accordingly (Fig.~\ref{fig:synthesis_pipeline} A). 
Table~\ref{tab:datasize} presents  segmentation results using only synthetic data for training the backbone. As the amount of synthetic data increases, the segmentation performance improves, particularly at the object level, highlighting the effectiveness of our synthetic dataset in enhancing segmentation performance.

\section{Discussion}
We propose a unified framework for microscopy image segmentation using shape and object regularized data synthesis but only using point-annotated images for training.  
As demonstrated via comprehensive validation and ablation analysis, our new framework outperforms existing baselines and obtains results similar to models trained on real images with dense annotations while significantly reducing annotation efforts.
Our current work leaves some open questions that will be addressed in future research. First, we currently perform offline data synthesis prior to training the segmentation model. Potential benefits of generating data online needs further investigation.
Additionally, we plan to incorporate additional prior knowledge, such as accounting for appearance variations among different tissue types, to generate more complex and anatomically-diverse objects. 

\section{Data use declaration and acknowledgment}
A retrospective analysis was performed on one open-access
datasets \cite{kumar2017dataset,kumar2019multi}, all
of whom complied with the Declarations of Helsinki and
received approval at their respective institutions. This work is supported by the grants NIH NIBIB R01EB021391, NIH 1R01EY030770-01A1, NIH-NEI 2R01EY013178-15,  and the New York Center for Advanced Technology in Telecommunications (CATT).

\newpage
\bibliographystyle{splncs04}
\bibliography{reference}

\newpage
\section{Appendix}
\begin{figure}[h]
    \centering
    \includegraphics[width=\textwidth]{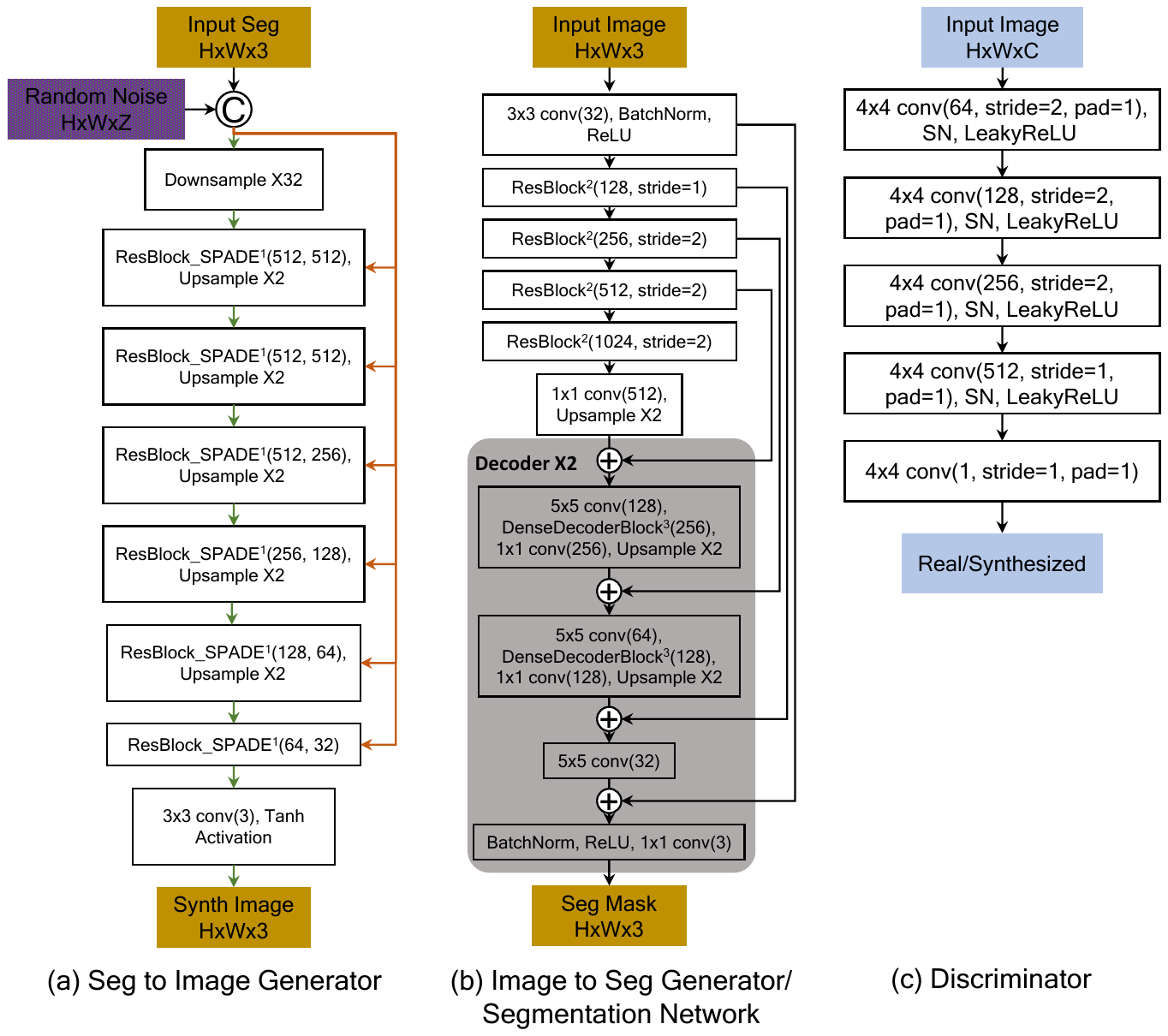}
    \caption{Detailed architectures of the (a) mask-to-image generator, (b) image-to-mask generator/ downstream segmentation network, and (c) the discriminator. 
    In (b), two decoder branches are used for the horizontal/vertical map and the binary segmentation map prediction, respectively. \texttt{ResBlock\_SPADE} follows[28], \texttt{ResBlock} and \texttt{DenseDecoderBlock} follow [11].\texttt{SN} indicates spectral normalization.}
    \label{fig:arch}
\end{figure}
\begin{table}[t]
\setlength{\tabcolsep}{1pt}
\centering
\begin{tabular}{ccccccc}
\toprule
\multirow{2}{*}{Config} & \multicolumn{1}{c}{\multirow{2}{*}{Randomness}} & \multicolumn{1}{c}{\multirow{2}{*}{Point Reg.}} & \multicolumn{2}{c}{Pixel level}                     & \multicolumn{2}{c}{Object level} \\ \cmidrule(lr){4-5} \cmidrule(lr){6-7}
 & \multicolumn{1}{c}{} & \multicolumn{1}{c}{} 
 & IoU $\uparrow$ & F1 $\uparrow$ 
 & Dice $\uparrow$ & AJI $\uparrow$ \\ \midrule
(a) & - & - & 0.483(0.19)    & 0.622(0.23) & 0.601(0.20)     & 0.413(0.17)    \\
(b) & \checkmark & - & 0.001(0.0008) & \multicolumn{1}{l}{0.003(0.0016)} & 0.003(0.0015)  & 0.001(0.0007) \\
(c) & \checkmark & \checkmark & 0.636(0.05)    & 0.776(0.04) & 0.745(0.04)     & 0.569(0.05)   \\ \bottomrule
\end{tabular}
\label{tab:additional_results}
\caption{Ablation comparison of downstream segmentation using data generated from different configurations. (a) is a baseline CycleGAN model; (b) injects randomness, which reduces inception distances and increases diversity (see main text Table.1). However, segmentation collapses due to erroneous object manipulation (see Fig.2(b)). Our final model (c) achieves the best segmentation results.}
\end{table}

\begin{figure}[h]
    \centering
    \includegraphics[width=\textwidth]{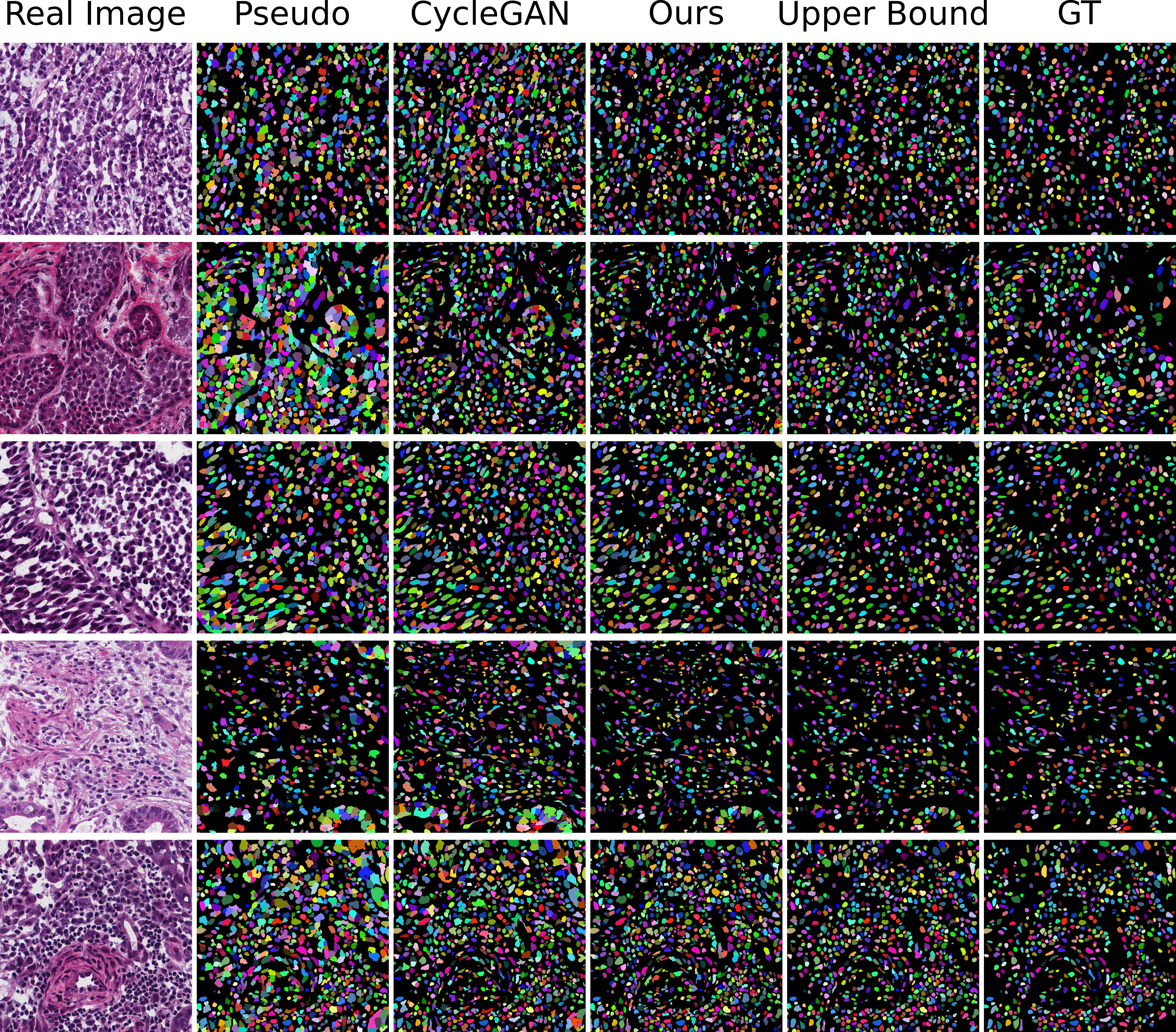}
    \caption{More results to supplement Fig. 3 in the main text.}
    \label{fig:arch}
\end{figure}
\begin{table}[h]
\setlength{\tabcolsep}{1pt}
\centering
\begin{tabular}{l|l|ll}
\toprule
\multirow{2}{*}{}   & \multicolumn{1}{c|}{\multirow{2}{*}{Mask-to-image GAN}} & \multicolumn{2}{c}{Segmentation}                                                                                                                                  \\ \cline{3-4} 
                    & \multicolumn{1}{c|}{}                                   & \multicolumn{1}{l|}{\begin{tabular}[c]{@{}l@{}}phase1: \\ train decoder\end{tabular}} & \begin{tabular}[c]{@{}l@{}}phase2: \\ train encoder\&decoder\end{tabular} \\ 
                    \midrule
init. learning rate & $2\times10^{-4}$ for G \&D                              & $1\times10^{-3}$                                                       \\ \hline
learning rate decay & linear (for 1000 epochs)                                & \multicolumn{2}{l}{step($1\times10^{-4}$ after 25 epochs)}                                                                                                         \\ \hline
optimizer           & Adam($\beta_1=0.5$, $\beta_2=0.999$)                    & \multicolumn{2}{l}{Adam($\beta_1=0.9$, $\beta_2=0.999$)}                                                                                                          \\ \hline
batch size          & 8                                                       & \multicolumn{1}{l|}{8}                                                                & 16                                                                        \\ \hline
crop size           & 256                                                     & \multicolumn{2}{l}{256}                                                                                                                                           \\ \hline
augmentation        & random flip                                             & \multicolumn{2}{l}{random flip, color jit, blur, noise}                                                                                                           \\ \hline
train epochs        & 4000                                                    & \multicolumn{2}{l}{50}                                                                                                                                            \\ \bottomrule
\end{tabular}
\caption{Training details for the mask-to-image GAN and the ad-hoc segmentation. The training of the segmentation network is two-stage following [11].}
\end{table}

\end{document}